\documentclass{article}

\usepackage{arxiv}

\usepackage[utf8]{inputenc} 
\usepackage[T1]{fontenc}    
\usepackage{hyperref}       
\usepackage{url}            
\usepackage{booktabs}       
\usepackage{amsfonts}       
\usepackage{nicefrac}       
\usepackage{microtype}      
\usepackage{lipsum}		
\usepackage{graphicx}
\usepackage[natbibapa]{apacite}
\usepackage{doi}
\usepackage{tabularx}
\usepackage{courier}  
\usepackage{listings}
\usepackage{xcolor}
\usepackage{longtable}

\title{Evaluation is all you need. Prompting Generative Large Language Models for Annotation Tasks in the Social Sciences. A Primer using Open Models}

\author{ \href{https://orcid.org/0000-0002-1174-449X}{\includegraphics[scale=0.06]{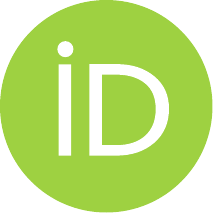}\hspace{1mm}Maximilian Weber}\thanks{This pre-print is currently under review. Please contact the first author via email with any feedback} \\
	Department of Social Science\\
	Goethe University\\
	Frankfurt, Germany \\
	\texttt{m.weber@soz.uni-frankfurt.de} \\
	\And
	{\hspace{1mm}Merle Reichardt} \\
	Department of Computer Science\\
	University Leipzig\\
	Leipzig, Germany \\
}

\renewcommand{\headeright}{Preprint}
\renewcommand{\undertitle}{Preprint}
\renewcommand{\shorttitle}{Prompting Generative Large Language Models for Annotation Tasks in the Social Sciences}

\hypersetup{
pdftitle={Prompting Generative Large Language Models for Annotation Tasks in the Social Sciences},
pdfsubject={q-bio.NC, q-bio.QM},
pdfauthor={Maximilian Weber, Merle Reichardt},
pdfkeywords={Open Models, LLM, Annotation},
}

\begin{document}
\maketitle

\begin{abstract}
	This paper explores the use of open generative Large Language Models (LLMs) for annotation tasks in the social sciences. The study highlights the challenges associated with proprietary models, such as limited reproducibility and privacy concerns, and advocates for the adoption of open (source) models that can be operated on independent devices. Two examples of annotation tasks, sentiment analysis in tweets and identification of leisure activities in childhood aspirational essays are provided. The study evaluates the performance of different prompting strategies and models (neural-chat-7b-v3-2, Starling-LM-7B-alpha, openchat\_3.5, zephyr-7b-alpha and zephyr-7b-beta). The results indicate the need for careful validation and tailored prompt engineering. The study highlights the advantages of open models for data privacy and reproducibility. 
\end{abstract}


\section{Introduction}
The advancement of Large Language Models (LLMs) has opened up new avenues for tackling annotation tasks in the field of social sciences. These models, especially the newer iterations like Chat-GPT or GPT-4, are now being used to annotate textual data \citep{Gilardi_Alizadeh_et_al._2023,Heseltine_Hohenberg_2023,Møller_Dalsgaard_et_al._2023,Ziems_Held_et_al._2023}, which can be helpful for analyzing various social and political phenomena \citep{Törnberg_2023,Ziems_Held_et_al._2023}. However, a significant challenge arises when there is a necessity to share research data with proprietary and closed models that are provided by companies due to the utilization of APIs \citep{Ollion_Shen_et_al._2023,Spirling_2023}. This is particularly concerning in scenarios where data sharing is not preferable due to data privacy. In light of this, open models which can be operated on independent devices like university servers, present a viable alternative \citep{Alizadeh_Kubli_et_al._2023}. They allow researchers to harness the potential of generative large language models without compromising data security. This paper endeavors to promote the adoption of open models by providing two examples and guidelines for leveraging them instead of proprietary models for annotation tasks within the social sciences.

\begin{figure}[ht]
    \centering
    \includegraphics[width=0.75\textwidth]{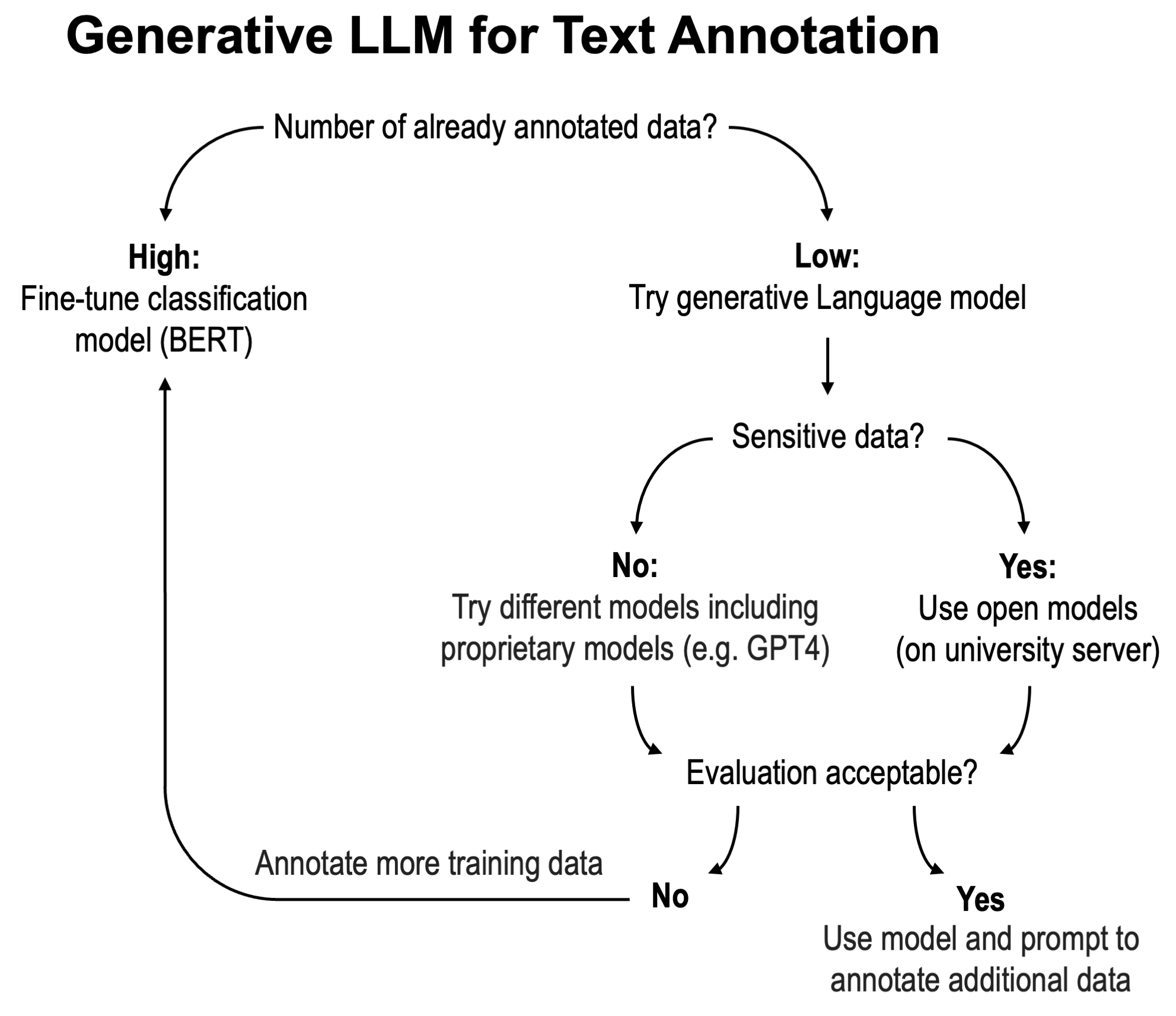}
    \caption{Decision tree for the use of generative LLM for text annotation}
    \label{fig:decision_tree}
\end{figure}

Researchers have identified several challenges associated with generative large language models, such as limited reproducibility, privacy and copyright concerns, and the prevalence of English-centric models \citep{Ollion_Shen_et_al._2023}. These issues indeed warrant attention, particularly in the context of proprietary models. However, the first two challenges can be mitigated through the implementation of open models that are available for local download and execution. In terms of the English language bias, we concur with the observation and anticipate the emergence of more open models that demonstrate satisfactory performance across diverse languages. We want to emphasize the need for carefully considering the use cases of generative language models. If they are used for annotation or classification tasks, the need for evaluation becomes important, as illustrated in Figure~\ref{fig:decision_tree}. The figure also illustrates a decision-making process for selecting an appropriate Large Language Model (LLM) for text annotation, emphasizing the consideration of data sensitivity and the quantity of annotated data available. This framework advocates for a strategic approach to model selection, wherein open models are preferred for sensitive data to enhance privacy and reproducibility, while proprietary models like GPT-4 are options for non-sensitive data. If these models fall short of expectations, it may be useful to fine-tune established text classification models like BERT, which necessitates the expansion of annotated training datasets.

This paper presents an exploration of five English models on two different annotation tasks.  With the rapid advancements in the field, a growing number of models are becoming available for various languages through a model database called HuggingFace, for example for Chinese (Qwen/Qwen-7B-Chat\footnote{\url{https://huggingface.co/Qwen/Qwen-7B-Chat}}), German (LeoLM/leo-hessianai-13b-chat\footnote{\url{https://huggingface.co/LeoLM/leo-hessianai-13b-chat}}) as well as multilingual models, for instance, Mixtral-8x7B-v0.1\footnote{\url{https://huggingface.co/mistralai/Mixtral-8x7B-v0.1}}. To discover more and newer models, one can browse the model catalog on the HuggingFace platform at (huggingface.co/models\footnote{\url{https://huggingface.co/models?pipeline_tag=text-generation}}). 

A critical observation to note is that, as of now, traditional fine-tuned text classification models have shown to outperform such generative language models that have not been fine-tuned on a specific task \citep{Plaza-del-Arco_Nozza_et_al._2023,Thalken_Stiglitz_Mimno_Wilkens_2023}. Traditional text classification models can utilize training data to fine-tune an existing LLM for a text classification task, while test data is employed to evaluate the model's performance. For a comprehensive understanding of applying these models in social sciences annotation tasks, the paper by \cite{Wankmüller_2022} is recommended. However, in situations where there is a lack of a large amount of annotated data or there is a need to obtain a quick overview, the potential of generative LLMs is worth investigating. Through this paper, we offer insights on leveraging open models for data annotation tasks and showcase two examples.

\section{Large language models for social science}
Generative Pre-trained Transformer (GPT) models, especially the latest iterations like GPT-4, have demonstrated versatility in social research applications. They have been deployed in various tasks such as sentiment analysis, text genre recognition, and identifying attitudes in tweets  \citep{Huang_Kwak_et_al._2023,Zhu_Zhang_et_al._2023}. However, challenges remain, particularly in processing languages where fewer models are available and in contexts where up-to-date information is critical. The research suggests that while GPT models excel in some tasks like sentiment analysis and political affiliation classification, their performance can be limited when dealing with new or evolving topics, as seen in the work of \cite{Zhu_Zhang_et_al._2023} on the Russian-Ukrainian conflict.

The effectiveness of generative language models in research is significantly influenced by prompt strategies and model configurations. Different approaches, such as Zero-Shot, One-Shot, Few-Shot, or Chain-of-Thought prompting, can yield varying results. \cite{Zhong_Ding_et_al._2023} highlight how different prompt strategies can enhance the models' performance, with methods like Few-Shot-Chain-of-Thought prompting showing notable improvements. Furthermore, studies like those by \cite{Gilardi_Alizadeh_et_al._2023} and \cite{Törnberg_2023} emphasize the importance of the model's temperature setting, with lower temperatures generally yielding more consistent and accurate results. 

The integration of GPT models in social research poses both opportunities and challenges. On one hand, they offer a powerful tool for analyzing large datasets, with capabilities surpassing manual methods in some instances, as shown in the studies by \cite{Gilardi_Alizadeh_et_al._2023} and \cite{Törnberg_2023}. On the other hand, their reliability and the need for careful validation, as noted by \cite{Reiss_2023}, indicate that while GPT models can be a valuable asset in social research, they should be used with caution. Researchers must be mindful of the limitations, particularly in terms of the models' training data and their ability to adhere strictly to guidelines in certain tasks.

\cite{Spirling_2023} and \cite{Ollion_Shen_et_al._2023} express concern over the reliance on proprietary Large Language Models (LLMs) due to their lack of transparency and potential ethical issues. \cite{Spirling_2023} advocates for the development and adoption of open-source LLMs through collaborative efforts, emphasizing the need for more funding and academic-led initiatives. 

The term "open models" can be understood as a continuum of accessibility. At one end of this spectrum, we find fully open-source models, where the training data, the scripts used for training, and the models themselves are freely available for download. At the other end, there are models that are open in terms of the model's availability for download and use on personal devices, but where the training data remains proprietary. This is in stark contrast to models like Chat-GPT, which, as of now, can only be operated through an account with the company OpenAI. Such a setup, where a model is accessible exclusively through an interface controlled by the owning organization, significantly differs from open models that permit downloading and independent operation.

In the realm of social sciences, some studies have employed open models for annotation tasks. \cite{Alizadeh_Kubli_et_al._2023} utilized HuggingChat (‘oasst-sft-6-llama-30b’ version) and Flan (various sizes) to showcase the capabilities of open models. Their findings suggest that open-source LLMs not only surpass human annotators from MTurk in performance but also yield results comparable to those of ChatGPT in certain tasks.

\section{Prompting strategies}
A prompt is a detailed instruction that is provided to a LLM to customize it and extend or refine its capabilities \citep{White_Fu_et_al._2023}. The prompt can influence or program the output of a LLM by providing specific rules and guidelines that set the context for the conversation and the desired output format \citep{White_Fu_et_al._2023}. It is thereby formulated in natural language and accordingly resembles a human conversation more than a programming process. \cite{White_Fu_et_al._2023} refer to prompt engineering as the means by which LLMs are programmed. They emphasize the great potential of prompt engineering and describe that the right prompt can not only improve the accuracy of the output, but also create entirely new interaction paradigms.

Typical elements of a prompt include a system message and a user message. If this prompt is given to a generative language model you will obtain a response that is called assistant message. The system message can contain a task definition, which outlines the specific task or type of response expected from the LLM. This is often followed by context setting, where relevant background information or specific details are provided in the system message to guide the model's response. In the user message you can give a text to classify and a classification question, if you want to categorize data.

Imagine that you possess a collection of tweets and your objective is to identify the sentiment. You might formulate your inquiry as follows: "What sentiment is expressed in this tweet?" and specify a response format, perhaps classifying sentiments as 'positive', 'negative', or 'neutral'. Furthermore, a system message like "You are an assistant for analyzing sentiments in text," could be utilized to define the task for the language model. Therefore, this input would comprise the system message, the tweet being analyzed, and the specific sentiment assessment question as an user message. While this example represents the simplest setup, you could also employ various prompting strategies, the most common of which we will explain in the following. 

\subsection{Zero-, One- and Few-Shot-Prompting}
Zero-, one- and few-shot-prompting is explained and tested in detail by Brown et al. (2020). According to them, zero-shot means that the prompt consists only of an instruction describing the task, but does not include any demonstrations. This method is particularly convenient, but it also presents the greatest challenge and is not suitable for all types of tasks. In some cases, a zero-shot task can be unnecessarily difficult, as it cannot be formulated clearly. However, some tasks might be suited for zero-shot prompts, for example translation tasks \cite{Brown_Mann_et_al._2020}. 

In contrast to this, a one-shot prompt includes exactly one demonstration of the task, Brown et al. explain. This is to convey the content or the desired format of the output in order to obtain a more accurate result. A demonstration of the task is always an example of a context and the associated completion. Accordingly, a one-shot prompt contains a task definition, then a demonstration consisting of a context and a completion, and finally a second context whose associated completion is to be generated by the model. The few-shot-prompt differs from the one-shot-prompt only in the number of demonstrations. \cite{Brown_Mann_et_al._2020} use between 10 and 100 demonstrations for their few-shot prompt tests . In the further literature no consistent number of demonstrations can be found. The different prompt settings can be understood as the degree to which the model relies on task-specific data. The test results of Brown et al. clearly show that the few-shot-prompting leads to an improvement of the results. Table~\ref{tab:prompt-strategies} shows example structures of various prompting strategies that can be used for annotation purposes.

\begin{table}[]
\centering
\begin{tabular}{l|ll}
\cline{1-2}
Prompting Strategie & Example Structure &  \\ \cline{1-2}
Zero-shot & \begin{tabular}[c]{@{}l@{}}\{"role": "system",  "content": "Text of System Prompt"\},\\ \{"role": "user",  "content": "(Text to classify) + classification question"\}\end{tabular} &  \\ \cline{1-2}
One-shot & \begin{tabular}[c]{@{}l@{}}\{"role": "system",  "content": "Text of System Prompt"\},\\ \{"role": "user",  "content": "(Example text) + classification question"\},\\ \{"role": "assistant", "content": "Example classification"\},\\ \{"role": "user",  "content": "(Text to classify) + classification question"\}\end{tabular} &  \\ \cline{1-2}
Few-shot & \begin{tabular}[c]{@{}l@{}}\{"role": "system",  "content": "Text of System Prompt"\},\\ \{"role": "user",  "content": "(Example text) + classification question"\},\\ \{"role": "assistant", "content": "Example classification"\},\\ \{"role": "user",  "content": "(Example text) + classification question"\},\\ \{"role": "assistant", "content": "Example classification"\},\\ … more examples \\ \{"role": "user",  "content": "(Text to classify) + classification question"\}\end{tabular} &  \\ \cline{1-2}
Chain-of-Thought & \begin{tabular}[c]{@{}l@{}}\{"role": "system",  "content": "Text of System Prompt"\},\\ \{"role": "user",  "content": "(Text to classify) + reasoning question"\},\\ \{"role": "assistant", "content": "Reasoning"\},\\ \{"role": "user",  "content": "Classification question"\}\end{tabular} &  \\ \cline{1-2}
\end{tabular}
\caption{Example structures of different prompting strategies for annotation purposes
}
\label{tab:prompt-strategies}
\end{table}

\subsection{Chain-of-Thought Prompting}
\cite{Wei_Wang_et_al._2023} discuss the chain-of-thought (CoT) prompt strategy, which consists of a series of intermediary reasoning steps leading to the final output. The authors describe that the thought chains mimic a step-by-step thought process as it occurs in humans when they solve a complicated reasoning task. They show that sufficiently large language models can be enabled to produce a coherent series of intermediate steps leading to correct answers. Few-shot prompting demonstrates to the language model how a task is broken down into intermediate steps and each step is solved individually to produce the final answer. The model is given a task, but unlike a simple zero, one, and few-shot prompt, the model is asked to first generate a chain of thoughts, and then output its final answer. For example, if the model is given the task, "If the child has 2 apples and gets 2 more apples, how many apples does the child have?", the model would not immediately respond with the solution, but would first generate a chain of thought, such as, "The child has 2 apples. It gets 2 apples in addition. There are 2 + 2 = 4 apples. The answer is 4". The results of Wei et al. (2023, pp. 8-9) show that the few-shot CoT prompting led to a clear improvement in performance compared to the standard few-shot prompting in a large number of the investigated tasks in the areas of arithmetic reasoning, commonsense reasoning and symbolic reasoning.

\cite{Kojima_Gu_et_al._2023} show that chain-of-thought prompting does not necessarily need to be associated with the few-shot strategy, but also performs well in the zero-shot application. The zero-shot CoT prompt is formed by the authors using a standard zero-shot prompt and simply adding the phrase "Let's think step by step" (cf. ibid. p. 2). Despite the simplistic prompt design, this generates a plausible reasoning path and the correct answer. Tasks where the standard zero-shot approach has failed can be solved by the zero-shot CoT, according to \cite{Kojima_Gu_et_al._2023} . Its performance is still below that of the few-shot CoT prompting, but compared to simple zero-shot prompts, it achieves significantly better values. The authors emphasize that the zero-shot CoT prompting has the advantages of being particularly uncomplicated, versatile and applicable to different argumentation tasks. According to them, few-shot CoT prompting is much more complex and very sensitive to task related prompting designs, which means that the sample questions have to be very close to the task questions in order to achieve good results. Fundamentally, the authors want to point out the 'hidden' zero-shot capability of LLMs and counter the premise that large language models are few-shot learners (cf. ibid. pp. 2-3).

\cite{He_Lin_et_al._2023} propose a two-step prompt strategy, which they call "explain-then-annotate" (cf. ibid. p. 2). Here, in the first step, the LLM is prompted to explain examples of annotations that have already been made. In the second step the explanations of the model are used to build a prompt, which contains the explanations of the mode (cf. ibid. pp. 2-3). This prompt is a few-shot CoT prompt, which was however supplemented by the authors by the fact that the explanations given in the prompt as an example were not written by humans, but by the model itself. The few-shot CoT prompt added by the authors is compared with the performance of zero-shot prompts and few-shot prompts on three different task types. For all task types, the few-shot CoT prompt performed best (cf. ibid. pp. 4-6). 

\subsection{Self-Consistency Prompting}
\cite{Wang_Wei_et_al._2023} propose a prompt strategy, which they call self-consistency. It is based on the fundamental idea that complex thinking tasks can usually be answered via several solution paths (cf. ibid. pp.1-2). The authors have the opinion that with increasing analysis effort for a task, also the quantity of the possible argumentation ways increases. They make use of this by running a few-shot CoT prompt not only once, but several times. This generates different solution paths and possibly also different results. According to the authors, the different answers are reduced to only one answer by majority voting. The most frequent answer is the final result. \cite{Wang_Wei_et_al._2023} describe that this procedure corresponds to the human experience, according to which a result is more trustworthy if several, differing solutions lead to it (cf. ibid. pp. 2-3). The results of Wang et al. show that self-consistency prompting can significantly improve the performance of a language mode (cf. ibid. p. 9). In practice, they recommend starting with a number of five to ten paths, in order to keep the costs, which they consider a disadvantage in the strategy, low.

\subsection{Prompt Pattern}
\cite{White_Fu_et_al._2023} present a catalog of formulation aids, which they call 'prompt patterns'. They represent a method for knowledge transfer and are intended to help solve conversational problems (cf. ibid. p. 1). Since they operate only at the level of instructional formulation in natural language, they are not classical prompt strategies, but patterns by which prompts can be structured. They can be applied in all task domains and are applicable independently of prompt strategies. White et al. relate their catalog strongly to communication with LLMs in the sense of a dialog, as an aid to programming tasks, but nevertheless, the patterns are generally valid. The persona pattern presented by the authors is particularly interesting. The language model is asked to take on a certain role with sentences like "Act like person X" or "Give answers like person X would give" (cf. ibid. pp. 7-8). This can for example be included in the system message. 

\section{Example Use Case}
We are showcasing two examples of annotation tasks on tweets and childhood aspirational essays to demonstrate the applicability of the prompting classification method, not just limited to various types of documents but also to diverse annotation tasks. Our demonstration specifically revolves around the annotation of sentiments in texts and extracting information from aspirational essays. However, it's important to note that the methodology employed here is not confined to such tasks. It has broader applications in other areas of classification within the social sciences, similar to previous research where various natural language processing methods were utilized. Examples of these applications include the classification of speeches, social media data, and chat group messages based on specific research questions.

The Python code utilized for this study is available on GitHub. For detailed guidance on replicating our procedures or accessing example files, please refer to the "Replication and Examples Files" section.

\subsection{Data}
We use two datasets to describe the approach and compare different models. First, we use a dataset originally created for an experiment, which is often referred to SemEval-2017 Task 4 - Subtask A \citep{Rosenthal_Farra_et_al._2017}. The dataset consists of English-language tweets with annotated sentiments. Our source was the "gold test data" available on a GitHub repository, encompassing over 12000 tweets. Given our objective to illustrate model prompting mechanisms rather than conduct full sample analysis, we limited our focus to a random subset of 500 tweets from the dataset. The distribution of the sentiment labels within this subsample is detailed in Table~\ref{tab:tweets}. Data from this SemEval-2017 Task 4 - Subtask A have been used by others to demonstrate the abilities of ChatGPT \citep{Zhu_Zhang_et_al._2023}. \cite{Zhu_Zhang_et_al._2023} use GPT3.5 and a zero-shot prompt to annotated 9895 tweets, they report an accuracy of 64.9\% and a weighted F1-score of 0.646.

\begin{table}[]
\centering
\begin{tabular}{lrr}
\hline
\textbf{Label} & \multicolumn{1}{l}{\textbf{Random Sample of 500 Tweets}} & \multicolumn{1}{l}{\textbf{Original Dataset}} \\ \hline
negative & 171 (34.20\%) & 3972 (32.33\%) \\
neutral & 227 (45.40\%) & 5937 (48.33\%) \\
positive & 102 (20.40\%) & 2375 (19.33\%) \\ \hline
 & 500 (100.00\%) & 12284 (100.00\%) \\ \hline
\end{tabular}
\caption{Distribution of the labels on the provide gold standard dataset of tweets
}
\label{tab:tweets}
\end{table}

Secondly, we utilize essays written by children at the age of 11 \citep{UKDA-SN-5790-2}. The data was collected as part of the National Child Development Study, a cohort study from the UK. Children were asked to write a brief essay about their future aspirations. The instructions for the essay were as follows: "Imagine you are now 25 years old. Write about the life you are leading, your interests, your home life, and your work at the age of 25. (You have 30 minutes to complete this task)" \citep{UKDA-SN-5790-2}. The data was manually transcribed and is accessible from the UK Data Archive\footnote{\url{https://beta.ukdataservice.ac.uk/datacatalogue/studies/study?id=5790}}. Some of the essays are annotated for different categories, such as mentions of a future partner, work, or leisure activities. Our focus is on the identification of planned leisure activities. If mentioned, it is labeled as "yes"; if not, it is labeled as "no". Out of 490 essays, annotations are present in the data set, and we concentrate on this data. The distribution of the childhood aspirational essays is detailed in Table~\ref{tab:essays}.

\begin{table}[]
\centering
\begin{tabular}{lrl}
\cline{1-2}
\textbf{Label} & \multicolumn{1}{l}{\textbf{Category distribution}} &  \\ \cline{1-2}
Yes (Leisure activity mentioned) & 337  (68.78\%) &  \\
No (Leisure activity not mentioned) & 153   (31.22\%) &  \\ \cline{1-2}
 &  490 (100.00\%) &  \\ \cline{1-2}
\end{tabular}
\caption{Distribution of the labels on the provide gold standard dataset for childhood aspirational essays
}
\label{tab:essays}
\end{table}

\subsection{Models}
For demonstration purposes we use five models: berkeley-nest/Starling-LM-7B-alpha\footnote{\url{https://huggingface.co/berkeley-nest/Starling-LM-7B-alpha}}, openchat/openchat\_3.5\footnote{\url{https://huggingface.co/openchat/openchat_3.5}}, Intel/neural-chat-7b-v3-2\footnote{\url{https://huggingface.co/Intel/neural-chat-7b-v3-2}}, HuggingFaceH4/zephyr-7b-beta\footnote{\url{https://huggingface.co/HuggingFaceH4/zephyr-7b-beta}}, HuggingFaceH4/zephyr-7b-alpha\footnote{\url{https://huggingface.co/HuggingFaceH4/zephyr-7b-alpha}}. They are all available from the hugging face platform. These models all have 7b parameters and were frequently downloaded in November 2023. We set the temperature value to 0.2, this value can be set from 0 to 1, while 1 gives more creative answers. Previous research argues that lower values might be better suited for annotation problems \citep{Chae_Davidson_2023,Gilardi_Alizadeh_et_al._2023}. Before we run each model we set a seed, so that the same model outcome is produced a second time when running a prompt on a model.

These models are trained with different prompting syntax in the training data. Table~\ref{tab:chat_templates} gives an overview of a simple prompt with a system and a user message, as well as an example of a prompt with an assistant message, the latter can be used to provide examples. If one tries different models with different prompting syntax it might be helpful to use a prompting templating function. We make use of the chat templating available in the transformers library. 

\begin{table}[ht]
\centering
\tiny
\begin{tabular}{l|ll}
\hline
\textbf{Model} &
  \multicolumn{1}{l|}{\textbf{Starling-7B}} &
  \textbf{Zephyr 7B Alpha} \\ \hline
\textbf{Pattern} &
  \multicolumn{1}{l|}{\begin{tabular}[c]{@{}l@{}}\textless{}s\textgreater{}GPT4 Correct System:\\ \{\{ system\_prompt \}\}\\ \textless{}|end\_of\_turn|\textgreater\\ GPT4 Correct User:\\ \{\{ user\_message \}\} \\ \textless{}|end\_of\_turn|\textgreater\\ GPT4 Correct Assistant:\end{tabular}} &
  \begin{tabular}[c]{@{}l@{}}\textless{}|system|\textgreater \\ \{\{ system\_prompt \}\}\\ \textless{}/s\textgreater\\ \textless{}|user|\textgreater\\ \{\{ user\_message \}\}\\ \textless{}/s\textgreater\\ \textless{}|assistant|\textgreater{}\end{tabular} \\ \hline
\textbf{Example} &
  \multicolumn{1}{l|}{\begin{tabular}[c]{@{}l@{}}\textless{}s\textgreater{}GPT4 Correct System:\\ Text of System Prompt\\ \textless{}|end\_of\_turn|\textgreater\\ GPT4 Correct User:\\ Question and text to classify\\ \textless{}|end\_of\_turn|\textgreater\\ GPT4 Correct Assistant:\end{tabular}} &
  \begin{tabular}[c]{@{}l@{}}\textless{}|system|\textgreater\\ Text of System Prompt \textless{}/s\textgreater\\ \textless{}|user|\textgreater\\ Question and text to classify\\ \textless{}/s\textgreater\\ \textless{}|assistant|\textgreater{}\end{tabular} \\ \hline
\textbf{Use of chat templating} &
  \multicolumn{2}{l}{\begin{tabular}[c]{@{}l@{}}\{"role": "system", "content":"Text of System Prompt"\},\\ \{"role": "user", "content": "Question and text to classify"\},\end{tabular}} \\ \hline
\textbf{Pattern with assistant message} &
  \multicolumn{1}{l|}{\begin{tabular}[c]{@{}l@{}}\textless{}s\textgreater{}GPT4 Correct System:\\ \{\{ system\_prompt \}\}\\ \textless{}|end\_of\_turn|\textgreater\\ GPT4 Correct User:\\ \{\{ user\_message \}\} \\ \textless{}|end\_of\_turn|\textgreater\\ GPT4 Correct Assistant:\\ \{\{ assistant\_message \}\}\\ \textless{}|end\_of\_turn|\textgreater\\ GPT4 Correct User:\\ \{\{ user\_message \}\} \textless{}|end\_of\_turn|\textgreater\\ GPT4 Correct Assistant:\end{tabular}} &
  \begin{tabular}[c]{@{}l@{}}\textless{}|system|\textgreater \\ \{\{ system\_prompt \}\}\\ \textless{}/s\textgreater\\ \textless{}|user|\textgreater\\ \{\{ user\_message \}\}\\ \textless{}/s\textgreater\\ \textless{}|assistant|\textgreater\\ \{\{ assistant\_message \}\}\\ \textless{}/s\textgreater\\ \textless{}|user|\textgreater\\ \{\{ user\_message \}\}\\ \textless{}/s\textgreater\\ \textless{}|assistant|\textgreater{}\end{tabular} \\ \hline
\textbf{Example} &
  \multicolumn{1}{l|}{\begin{tabular}[c]{@{}l@{}}\textless{}s\textgreater{}GPT4 Correct System:\\ Text of System Prompt\\ \textless{}|end\_of\_turn|\textgreater\\ GPT4 Correct User:\\ Question and text to classify\\ \textless{}|end\_of\_turn|\textgreater\\ GPT4 Correct Assistant:\\ classification\\ \textless{}|end\_of\_turn|\textgreater\\ GPT4 Correct User:\\ Question and another text to classify \\ \textless{}|end\_of\_turn|\textgreater\\ GPT4 Correct Assistant:\end{tabular}} &
  \begin{tabular}[c]{@{}l@{}}\textless{}|system|\textgreater\\ Text of System Prompt \textless{}/s\textgreater\\ \textless{}|user|\textgreater\\ Question and text to classify \textless{}/s\textgreater\\ \textless{}|assistant|\textgreater\\ classification \textless{}/s\textgreater\\ \textless{}|user|\textgreater\\ Question and another text to classify\textless{}/s\textgreater\\ \textless{}|assistant|\textgreater{}\end{tabular} \\ \hline
\textbf{Use of chat\_templating} &
  \multicolumn{2}{l}{\begin{tabular}[c]{@{}l@{}}\{"role": "system",  "content": "Text of System Prompt"\},\\ \{"role": "user",  "content": "Question and text to classify"\},\\ \{"role": "assistant", "content": "classification"\},\\ \{"role": "user",  "content": "Question and another text to classify"\}\end{tabular}} \\ \hline
\end{tabular}
\caption{Chat templates for zero-shot and one-shot prompts}
\label{tab:chat_templates}
\end{table}

\section{Evaluation metrics}
To evaluate the performance of the models, two key strategies are used. The first uses the kappa statistic, which assesses the agreement between the decisions made by the model and the already annotated data, considering the possibility of random agreement. This step is vital in classification or annotation tasks as it ensures that the model's decisions are consistent and not just coincidental. The kappa score, which ranges from -1 to +1, is used to measure the reliability of agreement between two raters or sets of ratings. In interpreting the kappa score, a value between 0.8 and 1.0 means very good agreement, 0.6 to 0.8 indicates good agreement, 0.4 to 0.6 shows moderate agreement, 0.2 to 0.4 signifies fair agreement, and a value below 0.2 suggests poor agreement. By using these evaluation methods together, a comprehensive understanding of how well different prompting strategies accurately categorize sentiments with various open generative large language models can be obtained.

The second strategy is the calculation of precision, recall, and the F1-Score. Simply put, precision looks at how many of the items identified are relevant. Recall checks how many relevant items are identified. The F1-Score combines both precision and recall to give a single measure for accuracy and completeness. Also, the average F1-Score is calculated, which is the mean of the F1-Scores obtained for each class, giving an overall performance metric without considering class distribution. The weighted F1-Score is also calculated, where the F1-Score for each class is weighted by its relative frequency in the dataset. This accounts for class imbalance by giving more weight to classes with more instances. This is particularly important in datasets where some classes are more represented than others, making sure that the model's performance reflects its ability to accurately identify each category, proportional to its occurrence.

\section{Results} 
\begin{figure}[ht]
    \centering
    \includegraphics[width=\textwidth]{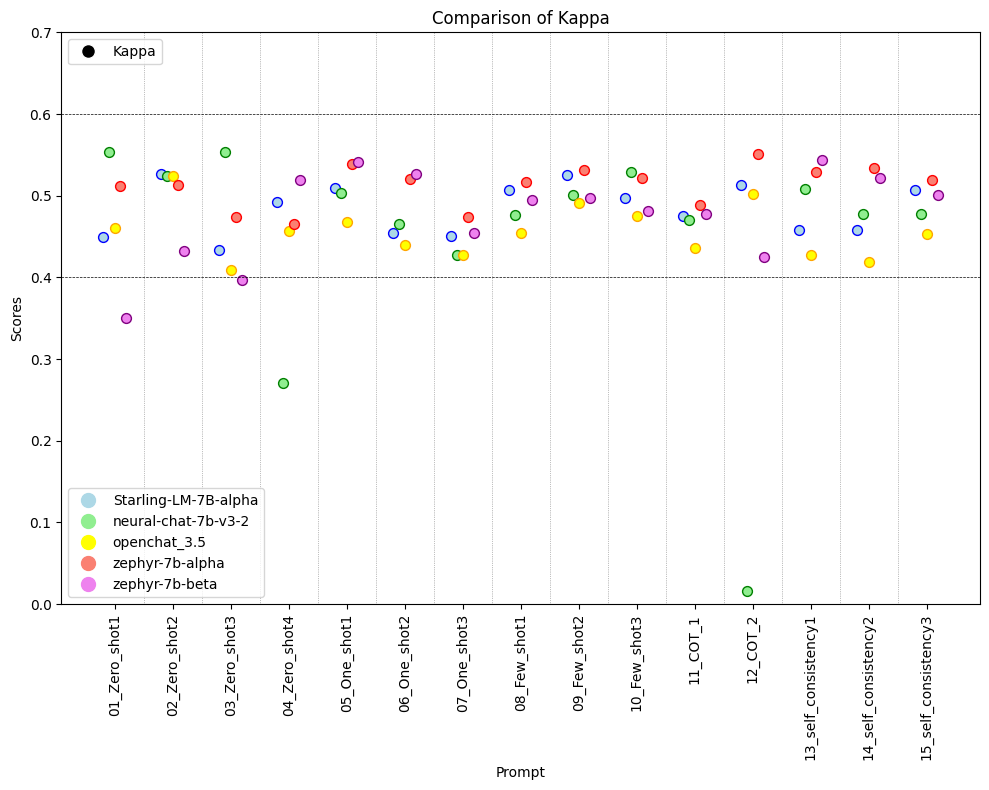}
    \caption{Evaluation results for the sentiment annotation for tweets. The figure displays 75 different prediction approaches, which include 5 models with 15 prompts each}
    \label{fig:myfigure_tweet_kappa}
\end{figure}

For both classification questions, we ran 15 prompts with five different models. The results for sentiment annotation are presented in Figure~\ref{fig:myfigure_tweet_kappa} with the kappa metric. A table for accuracy and macro F1-Score can be found in the appendix (Table~\ref{tab:tweets_models}). 

In the figure, it is clear that nearly all prompts and models have a moderate agreement with the gold data provided. Out of the 75 predictions, 71 fall within the moderate agreement level with a kappa score between 0.4 and 0.6. The best-performing prompts are observed for the zero-shot (1 and 2) prompting strategy with the neural-chat-7b-v3-2 model. Followed by zephyr-7b-alpha with Chain-of-Thought prompting. The kappa value for all these approaches is 0.55 and we observe an accuracy of 71 Percent (Table~\ref{tab:tweets_models}). Interestingly, the zephyr-7b-alpha model performs best for most of the used prompts, doing so 7 times out of 15 (Table~\ref{tab:tweets_models}). 

\begin{table}[]
\centering
\begin{tabular}{llllr}
\hline
 & \multicolumn{1}{c}{\textbf{Negative}} & \multicolumn{1}{c}{\textbf{Neutral}} & \multicolumn{1}{c}{\textbf{Positive}} & \multicolumn{1}{c}{\textbf{Overall}} \\ \hline
Precision & \multicolumn{1}{r}{0.78} & \multicolumn{1}{r}{0.72} & \multicolumn{1}{r}{0.61} & 0.70 \\
Recall & \multicolumn{1}{r}{0.73} & \multicolumn{1}{r}{0.65} & \multicolumn{1}{r}{0.81} & 0.73 \\
F1-Score & \multicolumn{1}{r}{0.75} & \multicolumn{1}{r}{0.68} & \multicolumn{1}{r}{0.70} & 0.71 \\ \hline
N & \multicolumn{1}{r}{171} & \multicolumn{1}{r}{227} & \multicolumn{1}{r}{102} & 500 \\
Macro Average F1-Score &  &  &  & 0.71 \\
Weighted Avg F1-Score &  &  &  & 0.71 \\
Accuracy &  &  &  & 0.71 \\
Kappa &  &  &  & 0.55 \\ \hline
\end{tabular}
\caption{Classification report for the best performing prompt of the neural-chat-7b-v3-2 model for classifying the sentiment}
\label{tab:classifying-sentiment}
\end{table}

Table~\ref{tab:classifying-sentiment} presents the classification results of the best performing approach. Precision measures the accuracy of our approach in correctly identifying tweets of each sentiment category. We achieved a precision of 0.78 for 'Negative' sentiments, 0.72 for 'Neutral,' and 0.61 for 'Positive.' This indicates that this approach is particularly adept at classifying 'Negative' tweets, while 'Positive' tweets present a slightly greater challenge in achieving precision. Recall, on the other hand, assesses the ability to capture all relevant instances of each sentiment category. The approach achieved a recall of 0.73 for 'Negative,' 0.65 for 'Neutral,' and 0.81 for 'Positive.' This means that our model excels at recognizing 'Positive' tweets, even though it may sometimes classify other sentiments as 'Positive'. The F1-Score balances precision and recall, offering a comprehensive evaluation metric. This approach obtained an F1-Score of 0.75 for 'Negative,' 0.68 for 'Neutral,' and 0.70 for 'Positive,' resulting in an overall F1-Score of 0.71. Table~\ref{tab:prompts_tweets} provides an overview of all prompts used for the leisure activities annotation.

\begin{figure}[ht]
    \centering
    \includegraphics[width=\textwidth]{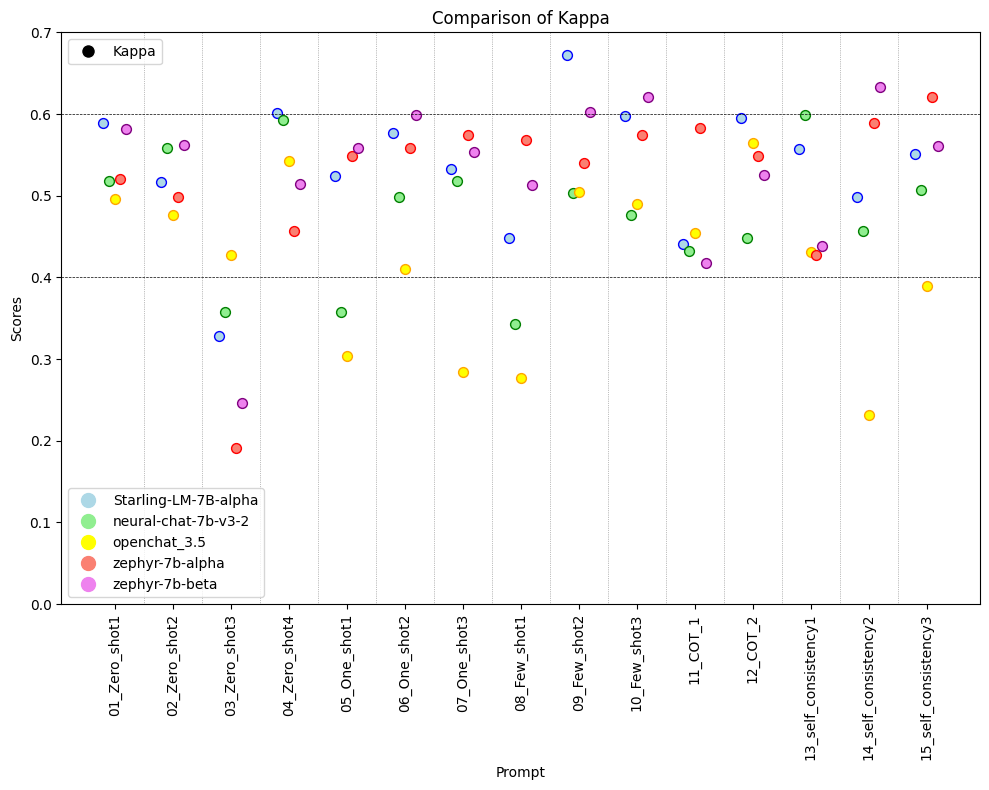}
    \caption{Evaluation results for the annotation indicating whether leisure activities are mentioned in childhood essays. The figure displays 75 different prediction approaches, which include 5 models with 15 prompts each}
    \label{fig:myfigure_plot_essay}
\end{figure}

For the annotation whether leisure activities are mentioned in childhood aspirational essays the results are more mixed, 64 annotation approaches out of 75 are above the 0.4 kappa value, indicating moderate agreement, 11 are below this threshold. However, we also observe 10 annotation approaches with kappa values above or equal the 0.6 threshold indicating a good agreement. The results for leisure activities annotation are presented in Figure~\ref{fig:myfigure_plot_essay} with the kappa metric. The best performing approach is observed for a few-shot prompt and the Starling-LM-7B-alpha model. The kappa value is 0.67 and we observe an accuracy of 0.86 and a macro F1-Score of 0.84 (Table~\ref{tab:modelsLeisure}). Against the observation for the sentiment approach, there is not one model that is most often outperforming the other models, we rather observe mixed findings (Table~\ref{tab:modelsLeisure}). For this annotation task, we observe good agreement values also for self-consistency prompting strategies, another few-shot approach and zero-shot with the zephyr-7b-alpha and zephyr-7b-beta models (Table~\ref{tab:modelsLeisure}).

\begin{table}[]
\centering
\begin{tabular}{lllr}
\hline
 & \multicolumn{1}{c}{\textbf{No}} & \multicolumn{1}{c}{\textbf{Yes}} & \multicolumn{1}{c}{\textbf{Overall}} \\ \hline
Precision & \multicolumn{1}{r}{0.82} & \multicolumn{1}{r}{0.88} & 0.85 \\
Recall & \multicolumn{1}{r}{0.73} & \multicolumn{1}{r}{0.93} & 0.83 \\
F1-Score & \multicolumn{1}{r}{0.77} & \multicolumn{1}{r}{0.90} & 0.84 \\ \hline
N & \multicolumn{1}{r}{153} & \multicolumn{1}{r}{337} & 490 \\
Macro Average F1-Score &  &  & 0.84 \\
Weighted Avg F1-Score &  &  & 0.86 \\
Accuracy &  &  & 0.86 \\
Kappa &  &  & 0.67 \\ \hline
\end{tabular}
\caption{Classification report for the best performing prompt of the Starling-LM-7B-alpha model for the annotation indicating whether leisure activities are mentioned in childhood essays}
\label{tab:classification-essays}
\end{table}

The Table~\ref{tab:classification-essays} specifically for the Starling-LM-7B-alpha model's best-performing prompt reveals important insights. With precision scores of 0.82 for 'No' and 0.88 for 'Yes,' the model shows a high degree of accuracy in its classifications. The model's recall for 'No' is 0.73, meaning it correctly identifies 73\% of the cases where leisure activities are not mentioned. In contrast, the recall for 'Yes' is notably higher at 0.93, indicating that the model is more effective at identifying essays that do mention leisure activities, capturing 93\% of such instances. This difference in recall suggests that the model is better tuned to recognize the presence of leisure activities than its absence. Overall, the higher recall for 'Yes' coupled with strong precision across both categories underscores the model's proficiency in detecting leisure activity mentions in essays. However, it also hints at potential areas for improvement in recognizing essays that do not mention such activities. Table~\ref{tab:prompts_childhood_aspirational_essays} provides an overview of all prompts used for the annotation of leisure activities.

\section{Discussion}
This study explored the application of open generative LLMs in the annotation of textual data, specifically focusing on sentiment analysis in tweets and the identification of leisure activities in childhood aspirational essays. The results reveal a nuanced performance, underscoring the critical need for careful validation and tailored approaches to prompt engineering.

In the sentiment analysis of tweets, we observed a general trend of moderate agreement across the majority of the prompts and models tested. Specifically, 71 out of 75 predictions fell within the moderate agreement range, as indicated by kappa scores between 0.4 and 0.6. This outcome suggests a consistent level of performance across different prompting strategies and models, yet it also highlights the room for improvement in achieving higher levels of agreement with the gold standard dataset. 

In contrast, the analysis of leisure activities in childhood aspirational essays yielded more varied results. While the majority of approaches still fell within the moderate agreement category, there was a notable presence of higher kappa values, indicating good agreement. This variance suggests that the nature of the text and the annotation task can significantly influence the effectiveness of prompting strategies. Unlike the sentiment analysis, there was no single model that often outperformed others.

These findings underscore the importance of considering both the nature of the annotation task and the characteristics of the models when designing prompts. The study demonstrates that while certain prompting strategies and models may be more effective in certain contexts, a one-size-fits-all approach is not applicable. The variation in performance across different models and tasks highlights the need for tailored approaches to prompt engineering, taking into account the specific requirements of each task and the capabilities of each model.

If different prompting strategies and models are to be tested, the annotated data should be split into development and test sets. This allows for the identification of an effective model using the development set, which can then be evaluated on the test set to assess its performance.

Aligning with existing research, our study confirms that generative models cannot be used directly "out of the box" for precise annotation tasks \citep{Pangakis_Wolken_et_al._2023,Thalken_Stiglitz_Mimno_Wilkens_2023}. The acceptability of evaluation metrics and outcomes varies, suggesting that in cases where results are not satisfactory, additional annotation of data and training of traditional text classification models may be necessary. This aligns with the recommendations of \cite{Pangakis_Wolken_et_al._2023} on the need of validation. 

The study highlights the advantage of open models for text annotation, particularly in scenarios where data privacy is crucial. 
These models can be downloaded and stored locally, providing a secure solution for handling sensitive data. 
This feature is especially important in research environments where external data sharing is not viable or desired, ensuring data confidentiality is maintained.
Furthermore, the ability to locally save and potentially reshare these models, combined with the option to set a specific seed for operations, enhances the reproducibility of results. 
This aspect marks a significant advancement in ensuring consistent and reliable outcomes in research using LLMs.
While English-centric models are prevalent and offer a wide range of options, the expanding landscape of LLMs is gradually accommodating more languages. This expansion is crucial for the inclusive application of LLMs in diverse linguistic contexts.

In conclusion, while open generative language models hold potential for automating text annotation tasks in social sciences, their effective deployment requires meticulous validation. Future efforts should focus on refining prompt engineering strategies, expanding language model availability, and continually fine-tuning models to suit specific annotation needs. 
This can be done with own instruction data sets. 
Moreover, generative models present a range of additional use cases beyond text annotation. 
These include image processing and the application in text-based tasks not just for annotation but also for extracting information or summarizing content \cite{Davidson_2023}. 
Examples of such applications include identifying the underlying reasons for specific annotations or for our essay example  determining the types of leisure activities mentioned in texts.
However, it's important to critically consider the energy consumption associated with operating these models. 
The significant energy requirements for using such models raise concerns about their environmental impact, an aspect that must be taken into account \citep{Luccioni_Jernite_et_al._2023}.

\section{Conclusion}
The main results of this study can be summarized across three key domains. Firstly, the study underscores the critical importance of validation of generative LLMs in text annotation tasks. It demonstrates that when results from generative LLMs are not satisfactory or don't meet the required accuracy, other approaches should be used, for example fine-tuning traditional text classification models should be considered as an alternative to improve evaluations.

Secondly, the research highlights the advantages of leveraging open models in scenarios where data privacy is a concern. These models provide a means to conduct research without the need to share sensitive data with external entities, ensuring data privacy is maintained.

Thirdly, the study points out the benefits of the ability to download and use models with a predefined seed. This feature ensures the reproducibility of results, which is essential in scientific research, particularly in fields where consistent results are crucial.

\section*{Replication and Examples Files}
After acceptance for publication we will make available all scripts needed for replication. Please refer to our github repository for any updates:  \href{https://github.com/textspur/prompting_open_llm}{https://github.com/textspur/prompting\_open\_llm}

\bibliographystyle{apacite}
\bibliography{references}  

\pagebreak 
\appendix
\section*{Appendix }
\label{appendix}
\subsection*{Additional tables}

\begin{longtable}[c]{llrrrr}
\hline
\textbf{Key} &
  \multicolumn{1}{c}{\textbf{Model}} &
  \multicolumn{1}{c}{\textbf{Kappa}} &
  \multicolumn{1}{c}{\textbf{Accuracy}} &
  \multicolumn{1}{c}{\textbf{Weighted F1-Score}} &
  \multicolumn{1}{c}{\textbf{Macro Avg F1-Score}} \\ \hline
\endhead
\hline
\endfoot
\endlastfoot
01\_Zero\_shot1        & neuralChat           & 0.55 & 0.71 & 0.71 & 0.71 \\
02\_Zero\_shot2        & Starling-LM-7B-alpha & 0.53 & 0.69 & 0.69 & 0.70 \\
03\_Zero\_shot3        & neuralChat           & 0.55 & 0.71 & 0.71 & 0.71 \\
04\_Zero\_shot4        & zephyr-7b-beta       & 0.52 & 0.70 & 0.70 & 0.70 \\
05\_One\_shot1         & zephyr-7b-beta       & 0.54 & 0.71 & 0.71 & 0.71 \\
06\_One\_shot2         & zephyr-7b-beta       & 0.53 & 0.69 & 0.69 & 0.69 \\
07\_One\_shot3         & zephyr-7b-alpha      & 0.47 & 0.64 & 0.63 & 0.64 \\
08\_Few\_shot1         & zephyr-7b-alpha      & 0.52 & 0.68 & 0.68 & 0.68 \\
09\_Few\_shot2         & zephyr-7b-alpha      & 0.53 & 0.70 & 0.70 & 0.70 \\
10\_Few\_shot3         & neuralChat           & 0.53 & 0.69 & 0.69 & 0.69 \\
11\_COT\_1             & zephyr-7b-alpha      & 0.49 & 0.66 & 0.65 & 0.66 \\
12\_COT\_2             & zephyr-7b-alpha      & 0.55 & 0.71 & 0.71 & 0.70 \\
13\_self\_consistency1 & zephyr-7b-beta       & 0.54 & 0.71 & 0.71 & 0.71 \\
14\_self\_consistency2 & zephyr-7b-alpha      & 0.53 & 0.69 & 0.69 & 0.69 \\
15\_self\_consistency3 & zephyr-7b-alpha      & 0.52 & 0.68 & 0.68 & 0.69 \\ \hline
\caption{ The best-performing model for each tested prompt in classifying the sentiment of tweets}
\label{tab:tweets_models}\\
\end{longtable}

\begin{longtable}[c]{llrrrr}
\hline
\textbf{Key} &
  \multicolumn{1}{c}{\textbf{Model}} &
  \multicolumn{1}{c}{\textbf{Kappa}} &
  \multicolumn{1}{c}{\textbf{Accuracy}} &
  \multicolumn{1}{c}{\textbf{Weighted F1-Score}} &
  \multicolumn{1}{c}{\textbf{Macro Avg F1-Score}} \\ \hline
\endhead
\hline
\endfoot
\endlastfoot
01\_Zero\_shot1        & Starling-LM-7B-alpha & 0.59 & 0.84 & 0.83 & 0.79 \\
02\_Zero\_shot2        & zephyr-7b-beta       & 0.56 & 0.80 & 0.81 & 0.52 \\
03\_Zero\_shot3        & openchat             & 0.43 & 0.77 & 0.76 & 0.71 \\
04\_Zero\_shot4        & Starling-LM-7B-alpha & 0.60 & 0.84 & 0.83 & 0.80 \\
05\_One\_shot1         & zephyr-7b-beta       & 0.56 & 0.83 & 0.82 & 0.52 \\
06\_One\_shot2         & zephyr-7b-beta       & 0.60 & 0.83 & 0.83 & 0.80 \\
07\_One\_shot3         & zephyr-7b-alpha      & 0.57 & 0.82 & 0.82 & 0.79 \\
08\_Few\_shot1         & zephyr-7b-alpha      & 0.57 & 0.83 & 0.82 & 0.78 \\
09\_Few\_shot2         & Starling-LM-7B-alpha & 0.67 & 0.86 & 0.86 & 0.84 \\
10\_Few\_shot3         & zephyr-7b-beta       & 0.62 & 0.83 & 0.84 & 0.54 \\
11\_COT\_1             & zephyr-7b-alpha      & 0.58 & 0.82 & 0.82 & 0.53 \\
12\_COT\_2             & Starling-LM-7B-alpha & 0.60 & 0.84 & 0.83 & 0.80 \\
13\_self\_consistency1 & neuralChat           & 0.60 & 0.83 & 0.83 & 0.80 \\
14\_self\_consistency2 & zephyr-7b-beta       & 0.63 & 0.84 & 0.84 & 0.82 \\
15\_self\_consistency3 & zephyr-7b-alpha      & 0.62 & 0.83 & 0.84 & 0.81 \\ \hline
\caption{The best-performing model for each tested prompt in classifying the mention of leisure activities}
\label{tab:modelsLeisure}\\
\end{longtable}

\subsection*{Used Prompts}

%
\begin{longtable}{|p{2cm}|p{2cm}|p{3cm}|p{3cm}|p{5cm}|}
\hline
\textbf{Prompt Name} &
  \textbf{\textbf{Classification Type}} &
  \textbf{\textbf{System Prompt}} &
  \textbf{\textbf{Question}} &
  \textbf{Examples Used} \\ \hline
\endhead
01\_zeroshot1 & Zero-Shot & You are an assistant for classifying tweets for their sentiment. & Is the sentiment for this tweet positive, negative or neutral? Respond only with: ['positive', 'neutral', 'negative']? & None \\
\hline
02\_zeroshot2 & Zero-Shot & You are a reliable chat assistant for classifying tweets for their sentiment. & Is the sentiment for this tweet positive, negative or neutral? Respond with: \textbackslash n 'positive' \textbackslash n 'neutral' \textbackslash n 'negative' \textbackslash n & None \\
\hline
03\_zeroshot3 & Zero-Shot & Please help me to classify tweets for their sentiment. & Question: Is the sentiment for this tweet positive, negative or neutral (Respond only with: ['positive', 'neutral', 'negative'])? & None \\
\hline
04\_zeroshot4 & Zero-Shot & Please help me to classify tweets for their sentiment. & Question: Is the sentiment for this tweet positive, negative or neutral. Do not reply using a complete sentence, respond with only one word: ['positive', 'neutral', 'negative'] & None \\
\hline
05\_One\_shot1 & One-Shot & Please help me to classify tweets for their sentiment. & Is the sentiment for this tweet positive, negative or neutral (only respond with ['positive', 'neutral', 'negative'])? & [(" Want to start reading Paper Towns tomorrow. My sister told me the movie was awesome! ", "positive")] \\
\hline
06\_One\_shot2 & One-Shot & Please help me to classify tweets for their sentiment. & Is the sentiment for this tweet positive, negative or neutral (only respond with ['positive', 'neutral', 'negative'])? & [("Missing Foo Fighters concert on the 12th at the gorge though so that's a major bummer", "negative")] \\
\hline
07\_One\_shot3 & One-Shot & Please help me to classify tweets for their sentiment. & Is the sentiment for this tweet positive, negative or neutral (only respond with ['positive', 'neutral', 'negative'])? & [("Is anyone going to see Shawn in NYC tomorrow", "neutral")] \\
\hline
08\_Few\_shot1 & Few-Shot & Please help me to classify tweets for their sentiment. & Is the sentiment for this tweet positive, negative or neutral (only respond with ['positive', 'neutral', 'negative'])? & [("Want to start reading Paper Towns tomorrow. My sister told me the movie was awesome! ", "positive"), ("Missing Foo Fighters concert on the 12th at the gorge though so that's a major bummer", "negative"), ("Is anyone going to see Shawn in NYC tomorrow", "neutral")] \\
\hline
09\_Few\_shot2 & Few-Shot & Please help me to classify tweets for their sentiment. & Is the sentiment for this tweet positive, negative or neutral (only respond with ['positive', 'neutral', 'negative'])? & [("Just finished a great workout session, feeling energized and ready for the day!", "positive"), ("Can't believe my vacation is over. Back to the daily grind tomorrow.", "negative"), ("Thinking about what to cook for dinner tonight. Maybe pasta?", "neutral"), ("Absolutely loved the new art exhibit at the museum. So inspiring!", "positive"), ("Feeling under the weather today, hope it's not the flu.", "negative"), ("Anyone know a good place for coffee in downtown?", "neutral")] \\
\hline
10\_Few\_shot3 & Few-Shot & Please help me to classify tweets for their sentiment. & Is the sentiment for this tweet positive, negative or neutral (only respond with ['positive', 'neutral', 'negative'])? & [("Really enjoyed the new coffee shop on Main Street, the ambiance was perfect!", "positive"), ("Feeling frustrated with the constant roadworks on my daily commute.", "negative"), ("Looks like it's going to be another ordinary day at the office.", "neutral")] \\
\hline
11\_COT & Chain of Thought & Please help me to classify tweets for their sentiment. & Is the sentiment for this tweet positive, negative or neutral. Think about it and provide some thoughts: / --- Now choose one of the following options: Is the sentiment for the tweet positive, negative or neutral take your thoughts into account and respond with only one word 'positive', 'neutral' or 'negative'. One-word reply: & None \\
\hline
12\_COT & Chain of Thought & Please help me to classify tweets for their sentiment. & Is the sentiment for this tweet positive, negative or neutral. Think about it and provide some thoughts: / --- Now choose one of the following options: 'positive','neutral' or 'negative'. One-word reply: & [
        {
            "user": "I had a fantastic day at the beach with my family!",
            "assistant": "This tweet expresses happiness and enjoyment, indicating a positive sentiment about spending time at the beach.",
            "annotation": "positive"
        },
        {
            "user": "Feeling down today. Sometimes things just don't go as planned.",
            "assistant": "The tweet mentions feelings of being 'down' and disappointment. This suggests a negative sentiment due to things not going as planned.",
            "annotation": "negative"
        },
        {
            "user": "Just another regular day at work. Nothing special.",
            "assistant": "The tweet describes a routine day without any specific positive or negative emotions. This seems like a neutral sentiment.",
            "annotation": "neutral"
        },
        {
            "user": "Can't believe how bad the traffic was today. Totally ruined my mood.",
            "assistant": "The tweet talks about frustration with bad traffic and a ruined mood, which clearly indicates a negative sentiment.",
            "annotation": "negative"
        }
    ] \\
\hline
13\_self consistency1 & Self-Consistency & You are an assistant for classifying tweets for their sentiment. & Is the sentiment for this tweet positive, negative or neutral? Only respond with ['positive', 'neutral', 'negative'] : & None, with multiple iterations for consistency. \\
\hline
14\_self consistency2 & Self-Consistency & Please help me to classify tweets for their sentiment. & Is the sentiment for this tweet positive, negative or neutral (only respond with ['positive', 'neutral', 'negative'])? & [("Missing Foo Fighters concert on the 12th at the gorge though so that's a major bummer", "negative")], with multiple iterations for consistency. \\
\hline
15\_self consistency3 & Self-Consistency & Please help me to classify tweets for their sentiment. & Is the sentiment for this tweet positive, negative or neutral (only respond with ['positive', 'neutral', 'negative'])? & [("Want to start reading Paper Towns tomorrow. My sister told me the movie was awesome! ", "positive"), ("Missing Foo Fighters concert on the 12th at the gorge though so that's a major bummer", "negative"), ("Is anyone going to see Shawn in NYC tomorrow", "neutral")], with multiple iterations for consistency. \\
\hline
\caption{Prompts for classifying tweets for their sentiment }
\label{tab:prompts_tweets}\\
\end{longtable}

\begin{longtable}{|p{2cm}|p{2cm}|p{3cm}|p{3cm}|p{5cm}|}
\hline
\textbf{Prompt Name} &
  \textbf{\textbf{Classification Type}} &
  \textbf{\textbf{System Prompt}} &
  \textbf{\textbf{Question Format}} &
  \textbf{Examples Used} \\ \hline
\endhead
01\_zeroshot1 & Zero-Shot & You are an assistant for classifying childhood essays. & Are any leisure activities mentioned? Classify with 'yes' or 'no': & None \\
\hline
02\_zeroshot2 & Zero-Shot & You are a reliable assistant for classifying childhood essays. Is leisure activity or hobbies mentioned? & Is leisure activity or hobbies mentioned? Classify with 'yes' or 'no': & None \\
\hline
03\_zeroshot3 & Zero-Shot & Please help me to classify childhood essays. Are plans for free time mentioned? & Question: Are plans for free time mentioned? Classify with ['yes', 'no']: & None \\
\hline
04\_zeroshot4 & Zero-Shot & Please help me to classify childhood essays. Is leisure activity mentioned? & Question: Is leisure activity mentioned? Classify with 'yes' or 'no': & None \\
\hline
05\_One\_shot1 & One-Shot & Please help me to classify childhood essays. Is leisure activity mentioned? & Question: Is leisure activity mentioned? Classify with 'yes' or 'no': & [("I am 25 years I will work. Unlike my mother, I will not have children of mine.", "no")] \\
\hline
06\_One\_shot2 & One-Shot & Please help me to classify childhood essays. Is leisure activity mentioned? & Question: Is leisure activity mentioned? Classify with 'yes' or 'no': & [("I am 25 years I will work. Unlike my mother, I will not have children of mine. I will be rowing.", "yes")] \\
\hline
07\_One\_shot3 & One-Shot & I need assistance in analyzing childhood essays about the future. Determine if a leisure activity is mentioned. & Question: Does the essay mention a leisure activity? Answer with 'yes' or 'no': & [("My ambition is to become a doctor. I will study hard and dedicate my life to helping others.", "no")] \\
\hline
08\_Few\_shot1 & Few-Shot & Please help me to classify childhood essays about future aspirations. Determine if a future leisure activity is mentioned. Here are some examples to guide you: & --- End of essay. Is leisure activity mentioned? Answer with 'yes' or 'no': & [("I am 25 years I will work. Unlike my mother, I will not have children of mine.", "no"), ("My ambition is to become a doctor. I will study hard and dedicate my life to helping others.", "no")] \\
\hline
09\_Few\_shot2 & Few-Shot & Please help me to classify childhood essays about future aspirations. Determine if a future leisure activity is mentioned. Here are some examples to guide you: & --- End of essay. Is leisure activity mentioned? Answer with 'yes' or 'no': & [
        ("I am 25 years I will work. Unlike my mothe I will not have children of mine.", "no"),
        ("My ambition is to become a doctor. I will study hard and dedicate my life to helping others. ", "no"),
        ("My future objective is to become a software developer, specializing in artificial intelligence.", "no"),
        ("I dream of sailing across the ocean. I've always been fascinated by the sea and hope to own a boat one day to explore different coastal areas.", "yes"),
        ("I aspire to be an actor, performing on stages across the globe. In my free time, I enjoy attending theater workshops and watching plays.", "yes"),
        ("I'm excited about becoming a chef and experimenting with different cuisines. On weekends, I love cooking new recipes for my family and friends.", "yes"),
        ("One day, I hope to direct movies. I'm currently learning about film-making and scriptwriting in my spare time.", "yes")
    ] \\
\hline
10\_Few\_shot3 & Few-Shot & Please help me to classify childhood essays about future aspirations. Determine if a future leisure activity is mentioned. Here are some examples to guide you: & --- End of essay. Is leisure activity mentioned? Answer with 'yes' or 'no': & [
        ("I am 25 years I will work. Unlike my mothe I will not have children of mine.", "no"),
        ("My ambition is to become a doctor. I will study hard and dedicate my life to helping others. ", "no"),
        ("In my free time, I love to play basketball with my friends. I hope to continue playing basketball in college and possibly even join a professional league.", "yes"),
        ("I am determined to pursue a career in law, aiming to specialize in international law. I spend a lot of my work time reading about legal cases and preparing for debates.", "no")
    ] \\
\hline
11\_COT & Chain of Thought & Please help me to classify essays written by students at age 11. Is a leisure activity mentioned? & --- End of essay. Is a leisure activity mentioned? Think about it and provide some thoughts: & None \\
\hline
12\_COT & Chain of Thought & Please help me to classify essays written by students at age 11. Is a leisure activity mentioned? & --- End of essay. Is a leisure activity mentioned? Think about it and provide some thoughts: & [
        {
            "user": "In my future, I see myself studying at a prestigious university and becoming a scientist. Most of my time will be spent in research.",
            "assistant": "The essay talks about academic and career goals but does not mention any leisure activities like sports, hobbies, or relaxation.",
            "annotation": "no"
        },
        {
            "user": "When I grow up, I want to play professional football. I enjoy playing football every weekend with my friends.",
            "assistant": "This essay clearly mentions playing football, which is a leisure activity. The student talks about enjoying this sport in their free time.",
            "annotation": "yes"
        },
        {
            "user": "My dream is to help people as a doctor. In my free time, I would like to write novels, as writing is my passion.",
            "assistant": "The essay mentions two main aspects: a career goal of becoming a doctor and a leisure activity, which is writing novels.",
            "annotation": "yes"
        },
        {
            "user": "I aspire to be an engineer and invent new gadgets. I spend a lot of time reading about technology and creating small devices.",
            "assistant": "While the essay focuses on career aspirations in engineering, it does not explicitly mention any activities that are done for leisure.",
            "annotation": "no"
        }
    ]) \\
\hline
13\_self consistency1 & Self-Consistency & Please help me to classify essays written by students at age 11. & --- End of essay. Is leisure activity mentioned? Answer with 'yes' or 'no': & None \\
\hline
14\_self consistency2 & Self-Consistency & Please help me to classify essays written by students at age 11. & --- End of essay. Is leisure activity mentioned? Answer with 'yes' or 'no': & [("I am 25 years I will work. Unlike my mother, I will not have children of mine.", "no")] \\
\hline
15\_self consistency3 & Self-Consistency & Please help me to classify childhood essays about future aspirations. Determine if a future leisure activity is mentioned. Here are some examples to guide you: & --- End of essay. Is leisure activity mentioned? Answer with 'yes' or 'no': & [("In the future, I dream of becoming an astronaut and exploring space. On weekends, I plan to go hiking in the mountains.", "yes"), ("My goal is to study medicine and become a renowned surgeon, dedicating my life to saving lives and conducting medical research.", "no"),] \\
\hline
\caption{Prompts for classifying childhood aspirational essays for mentioning of leisure activities}
\label{tab:prompts_childhood_aspirational_essays}\\
\end{longtable}

\end{document}